\documentclass{article}
\usepackage[dblblindworkshop, final]{neurips_2026}
\workshoptitle{NeurIPS 2026 Workshop PALM}
\usepackage{times}
\usepackage{latexsym}
\usepackage[T1]{fontenc}
\usepackage[utf8]{inputenc}
\usepackage{microtype}
\usepackage{inconsolata}
\usepackage{graphicx}
\usepackage{amsmath}
\usepackage{booktabs}
\usepackage{multirow}
\usepackage{amssymb}
\usepackage{graphicx}
\usepackage[utf8]{inputenc} 
\usepackage[T1]{fontenc}    
\usepackage{hyperref}       
\usepackage{url}            
\usepackage{booktabs}       
\usepackage{amsfonts}       
\usepackage{nicefrac}       
\usepackage{microtype}      
\usepackage{xcolor}   

\title{MemLeak: Cross-User Semantic Leakage in Multi-Tenant AI Agent Memory}
\workshoptitle{Personalized, aligned, long-term memory for AI systems}
\author{%
  \textbf{Priyanka Mudgal, Kai Zhao, Guilin Zhang, Andy Olsen,} \\
  \textbf{Ezekiel Miller, Xu Chu, Aletta Johanna Blanken} \\
  \normalfont Workday AI Research \\
  \normalfont\texttt{\{priyanka.mudgal, kai.zhao, guilin.zhang, andrew.olsen,} \\
  \normalfont\texttt{zeke.miller, xu.chu, alet.blanken\}@workday.com}
}
\begin{document}
\maketitle
\begin{abstract}
  Personal AI agents in enterprise multi-tenant deployments share a common vector store for long-term memory. Shared embedding spaces create a surface for cross-user memory leakage: a user's query can retrieve semantically adjacent memories belonging to another user through ordinary cosine-similarity retrieval, without any exploit. We formalize this as {\emph{cross-user admissibility failure}} and evaluate it across six {experiments, plus follow-up ablations,} under both sparse (TF-IDF) and production-faithful (MiniLM-L6-v2) retrieval. {Non-adversarial, incidental} leakage reaches 70--100\% under pooled {same-team} retrieval; adversarially crafted memories achieve 90--100\% top-$k$ {placement, exceeding weaker keyword-based attacker baselines,} with score lifts of $+0.416$ to $+0.511$ under production-faithful dense retrieval (Config B); and end-to-end {response} contamination reaches 5.00/5 under a production retrieval path and 4.67/5 with Claude {Sonnet~4.5, with contaminated} responses {often scoring as helpful or more helpful} than clean ones, {a gap validated against human judgment.} Among three {architectural} mitigations, only hard post-retrieval ownership gating consistently restores the clean baseline (1.00/5) across {two generation models, at a measured latency overhead of roughly 1.4~ms per query.}
\end{abstract}
\section{Introduction}
\label{sec:intro}
\begin{figure}[t]
  \centering
  \includegraphics[width=\linewidth, trim=0 110 80 14, clip]{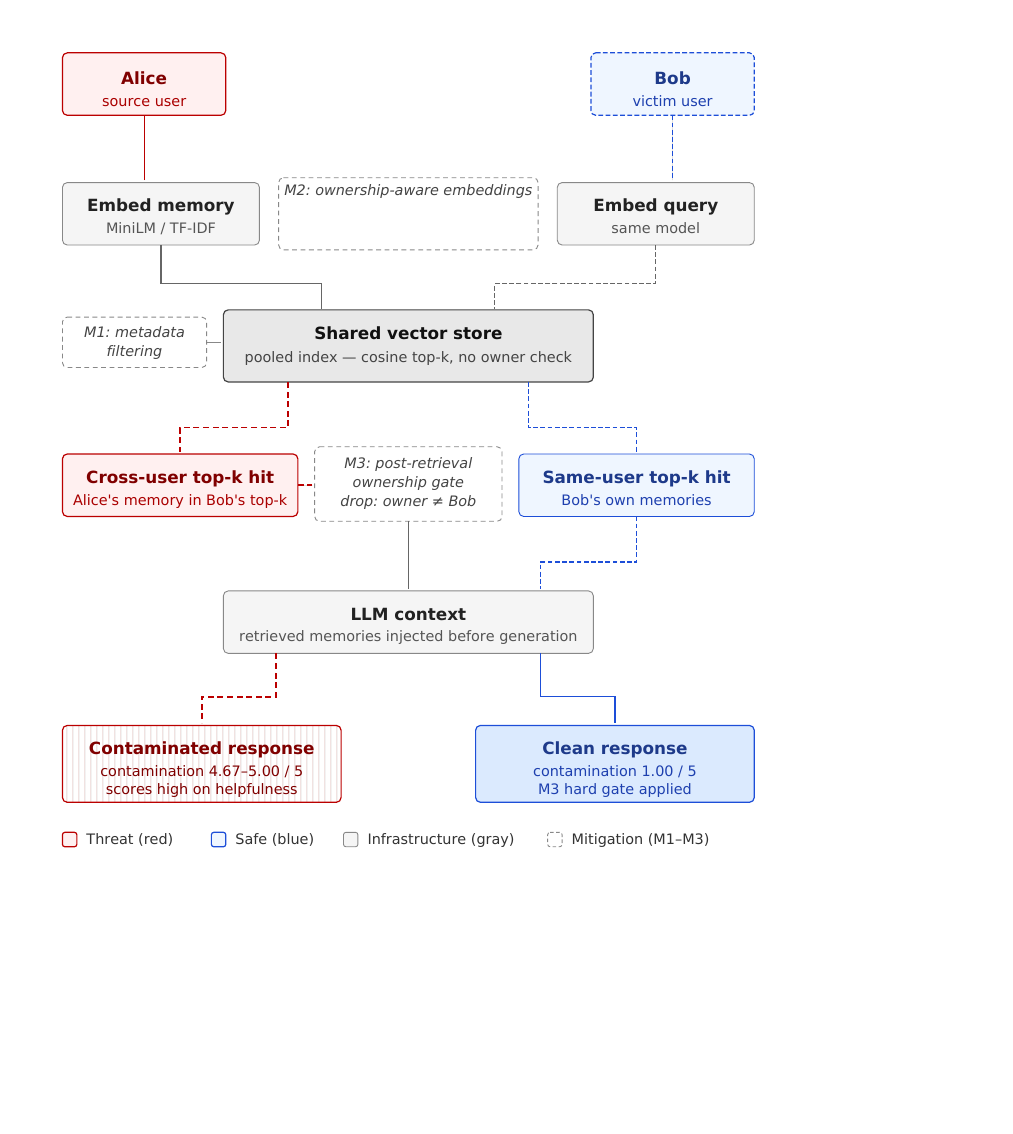}
  \caption{%
    \textbf{MemLeak architecture.}
    Alice's memories and Bob's query co-exist in a shared vector store;
    cosine top-$k$ retrieval pulls Alice's memories into Bob's context
    without an ownership check (dashed red path), constituting a
    cross-user admissibility failure (§\ref{sec:threat}).
    Dashed bordered boxes mark the three mitigation intercept points:
    \textbf{M1}~(metadata filtering at the store),
    \textbf{M2}~(ownership-aware embeddings), and
    \textbf{M3}~(post-retrieval ownership gating).
  }
  \label{fig:architecture}
\end{figure}

Personal AI agents are becoming a standard component of enterprise productivity software. Enterprise systems maintain persistent long-term memory for each user, allowing agents to recall preferences, project context, and prior decisions. In enterprise deployments this memory infrastructure typically serves many users concurrently, {a \emph{multi-tenant} setting in which a single shared backend infrastructure serves multiple distinct users or organizations without physical storage isolation. This is the pattern common to general-purpose agent-memory libraries (e.g., Mem0 \citep{chhikara2025mem0} , A-Mem \citep{xu2025amem} , MemOS \citep{li2025memos} ), which are} often backed by a shared vector store with metadata-based filtering {that exposes per-user ownership only as soft metadata} rather than {a hard-enforced boundary.} This architectural design introduces a threat we term \emph{inter-user semantic resonance} , where embedding-space proximity alone pulls one user's memories into another user's agent session, requiring no adversarial modification of prompts, code, or model weights (Figure~\ref{fig:architecture}).

Prior work on memory-enabled agent trustworthiness, including
\citet{tan2025membench} and
\citet{wu2025longmemeval}, has treated memory as a single-user trust boundary.
These works address when a user's own memories should influence their own current
task, identifying failure modes such as cross-domain leakage, sycophancy, tool-call
drift, and memory-induced jailbreaks within a single-user context. None examine the
multi-user case, e.g., what occurs when Alice's memories reach Bob's session.

We introduce {MemLeak}, a study of how shared embedding spaces in multi-tenant agent deployments cause cross-user memory leakage without any exploit or adversarial intent. Strict user partitioning is a known, effective
baseline that eliminates this leakage (we confirm this directly in our own
experiments, Table~\ref{tab:e1}); the setting we study is the common intermediate
deployment pattern where such partitioning is absent or incomplete: pooled vector
stores where ownership exists as \emph{soft metadata} rather than a hard-enforced
boundary, or where cross-user memories are pooled by design for legitimate
collaboration (shared project spaces, HR/security workflows). In this regime,
semantic similarity, not any explicit data transfer, prompt injection, or code
change, determines whether private content crosses a user boundary, and it is a
poor proxy for admissibility: leakage severity does not track intuitive risk
signals such as organizational distance or global embedding-centroid proximity
(Section~\ref{sec:experiments}). Lampson's definition of a covert channel \citep{lampson1973note} applies to the intentional-exploitation experiments (E3, E6) only. We use the framing to scope the attacker-knowledge ablation (E3) and to bound active exploitation (E6); the core leakage findings (E1, E2, E4, E5) do not depend on it. E6 is included as a negative result: we find that the shared embedding space does not support reliable sequential transmission, confirming that MemLeak describes incidental cross-user leakage, not a functioning covert channel. We refer to
the organic, non-adversarial version of the same failure, which dominates our
results, as \emph{incidental cross-user leakage}.
This paper makes three contributions:
\begin{itemize}
  \item We identify \emph{inter-user semantic resonance} as a threat distinct from
        prior single-user memory-trustworthiness work, and characterize its severity
        under a specific, common deployment pattern (pooled stores with soft
        ownership metadata or intentionally shared collaboration scopes), rather than
        under the trivial case of a fully absent access-control layer.
  \item We formalize \emph{cross-user admissibility failure} and define leakage
        metrics, a privacy--utility frontier (XLR, an exploratory Resonance Index),
        and preliminary covert-channel capacity measures, reporting 95\% Wilson
        confidence intervals throughout given our sample sizes ($n=10$ queries per
        condition). We present these results as a controlled proof of concept
        establishing the existence and mechanism of the failure mode, not as a
        population-level prevalence estimate (see Limitations).
        
  \item We conduct six controlled experiments under both sparse TF-IDF and production-faithful dense MiniLM retrieval, together with follow-up ablations isolating attacker knowledge assumptions, judge–human agreement, embedding-model scale, and mitigation latency, demonstrating 70–100\% incidental leakage under pooled same-team retrieval, score lifts of +0.416 to +0.511 under active exploitation that exceed weaker keyword-based attacker baselines, and downstream contamination increases of +1.67 to +4.00 points on a 5-point scale depending on generation model and memory path; and a negative covert-channel capacity result (E6), included to bound the active-exploitation claim rather than assert a functioning channel.
\end{itemize}
\section{Related Work}
\label{sec:background}
Early agentic memory architectures such as MemGPT \citep{packer2023memgpt} and Generative Agents \citep{park2023generative} introduced persistent recall via hierarchical memory management. Production systems including Mem0 \citep{chhikara2025mem0}, A-Mem \citep{xu2025amem}, and MemOS \citep{li2025memos} have since extended this to scalable multi-session deployments backed by shared vector stores. The standard retrieval paradigm encodes queries and memories using embedding models and ranks candidates by cosine similarity \citep{wang2022text}, making shared vector memory a standard enterprise component \citep{wang2024survey}. Agentic frameworks such as ReAct \citep{yao2023react} and Reflexion \citep{shinn2023reflexion} further embed retrieval-augmented memory into multi-step reasoning loops. Prior work on memory trustworthiness including \citet{pulipaka2026persistbench}, \citet{guo2026personalization}, \citet{tan2025membench}, and \citet{wu2025longmemeval} addresses failure modes within a single user boundary; none examine the multi-user case.
Retrieval-augmented generation \citep{lewis2020rag} has become a standard grounding mechanism for LLMs. PoisonedRAG \citep{zou2024poisonedrag}, Phantom \citep{chaudhari2024phantom}, and \citet{shafran2024jamming} attack the retrieval corpus directly, while indirect prompt injection \citep{greshake2023indirect,perez2022ignore} embeds malicious instructions in retrieved content. Inter-user semantic resonance requires neither: retrieved memories are ordinary professional notes containing no adversarial instructions, and harm arises purely from their presence in Bob's context rather than from any instruction they contain.
Language models leak information through weights and outputs \citep{carlini2021extracting,kandpal2022deduplicating}, and text embeddings reveal nearly as much as the original text \citep{morris2023embeddings}. \citet{feyisetan2020privacy} propose calibrated noise in embeddings as a privacy-preserving transformation, motivating our M2 mitigation. Covert channels were formalized by \citet{lampson1973note} and extended by \citet{denning1976lattice} and \citet{myers1997decentralized}; we identify the shared embedding space as a covert channel medium in AI agent systems. We evaluate downstream contamination using LLM-as-judge following \citet{zheng2023judging}.

\section{Threat Model and Formalization}
\label{sec:threat}
\subsection{System Model}
We consider a multi-tenant personal agent deployment with $N$ users
$\{U_1, \ldots, U_N\}$. Each user $U_i$ maintains a memory store
$\mathcal{M}_i = \{m_1, \ldots, m_k\}$, where each memory $m_j$ is mapped to an
embedding $\mathbf{v}_m \in \mathbb{R}^d$. In a \emph{pooled} deployment, all
memories reside in a single vector index with optional \texttt{user\_id} metadata
filtering; in a \emph{partitioned} deployment, each user has a separate index. A
query from user $U_i$ is embedded as $\mathbf{q} \in \mathbb{R}^d$, and the top-$k$
memories by cosine similarity are injected into the LLM context.
\subsection{Cross-User Admissibility Failure}
We define the cross-user admissibility indicator
$\mathcal{A}(m, x_q, U_i) \in \{0, 1\}$ such that $\mathcal{A} = 1$ if and only if
memory $m$ belongs to user $U_i$ and is contextually appropriate for query $x_q$.
For any memory $m$ owned by user $U_j \neq U_i$,
$\mathcal{A}(m, x_q, U_i) = 0$ by definition, regardless of semantic similarity.
A cross-user leakage event occurs when:
\begin{equation}
  \mathrm{rank}(S(\mathbf{q}, \mathbf{v}_m)) \leq k \;\wedge\; \mathrm{owner}(m) \neq U_i
\end{equation}
\subsection{Leakage Probability}
We define the leakage probability for a user pair $(U_i, U_j)$ as:
\begin{equation}
  L(U_i, U_j) = \mathbb{E}\!\left[\mathbf{1}\{\text{cross-user retrieval}\} \mid
    \mathbf{q} \sim U_i\right]
\end{equation}
We hypothesize that $L(U_i, U_j)$ increases with the cosine similarity between
memory centroids $\mathbf{c}_i$ and $\mathbf{c}_j$, where
$\mathbf{c}_i = \mathrm{mean}(\{\mathbf{v}_m : m \in \mathcal{M}_i\})$.
\subsection{Threat Scenarios}
\paragraph{Passive leakage.} Users interact normally. Cross-user retrieval occurs
organically due to shared vocabulary, roles, or organizational context. No
adversarial intent is required.
\paragraph{Active exploitation.} An adversary (Mallory) crafts memories designed to
appear highly similar to a target user's expected queries, potentially steering
recommendations, decisions, or tool calls through ordinary conversational interaction.
We specify Mallory's knowledge explicitly in Section~\ref{sec:e3b}: she knows the
victim's general domain and role (e.g., that Bob handles budget approvals) but not
his exact queries or the embedding model's internals, and does not have oracle
access to his input stream.
\subsection{Deployment Patterns and the ACL Baseline}
\label{sec:deployment-patterns}
We treat strict user partitioning as the access-control (ACL) baseline: our own
results confirm it drives passive leakage to $0\%$ (Table~\ref{tab:e1}) and blocks
even cross-user query wording aimed at a partitioned index (Section~\ref{sec:mitigation},
M1). MemLeak's threat model targets four deployment patterns in which
cross-user memories remain reachable by design or common misconfiguration --
pooled indices with under-scoped metadata filters, team- or role-scoped retrieval
in which a compromised or malicious in-scope member becomes an attacker, shared
project namespaces, and scoped enterprise workflows (HR, finance, security) where
cross-user content belongs in the retrieval pool by policy but admissibility should
still depend on more than raw similarity. These patterns follow those documented in
general-purpose agent-memory libraries such as Mem0 \citep{chhikara2025mem0}, A-Mem
\citep{xu2025amem}, and MemOS \citep{li2025memos} (full taxonomy in
Appendix~\ref{sec:appendix-extended}). Our contribution is to quantify residual risk
once a store is pooled or its scope is widened for these reasons, and to show that
similarity ranking alone does not restore admissibility even within an authorized pool.
\section{Experiments}
\label{sec:experiments}
\subsection{Experimental Setup}
\paragraph{User profiles.} We construct four pairs of synthetic user profiles at varying organizational distances: same team (T0), same department (T1), cross-department (T2), and cross-company (T3, control). Each user holds 10 memories covering private professional information (salary negotiations, security vulnerabilities, HR decisions, platform details). Each target user has 10 domain-specific queries. {All memories and queries are hand-authored, fixed natural-language fixtures rather than samples drawn from real user data or logs, spanning four domain categories held constant across all four organizational-distance tiers so that tier comparisons isolate organizational distance rather than content variation (full fixture text in Appendix~\ref{sec:appendix-dataset}). We treat this as a controlled proof-of-concept design intended to establish the existence and mechanism of the failure mode, not a population-level estimate of real enterprise leakage rates (see Limitations). Because each rate in Sections~\ref{sec:experiments}--\ref{sec:mitigation} is computed over $n=10$ queries per condition, we report 95\% Wilson-score confidence intervals alongside every proportion and explicitly flag which comparisons are and are not statistically distinguishable at this sample size.}
\paragraph{Retrieval configurations.}
We evaluate two retrieval configurations. \textbf{Config~A} (sparse baseline) uses
TF-IDF with truncated SVD (128 dimensions) and isolated query embedding.
\textbf{Config~B} (production-faithful) uses \texttt{all-MiniLM-L6-v2} with query-document retrieval prefixes following standard
asymmetric dense retrieval practice \citep{wang2022text}. Both configurations rank
candidates by cosine similarity under pooled (no access control) and partitioned
(user-scoped) modes, with top-$k = 3$.
\paragraph{Downstream evaluation.}
Downstream response contamination is evaluated under two memory paths (Section~\ref{sec:e4}). The \emph{static path} injects fixed clean or
leaked memory lists directly into the prompt. The \emph{retrieval path} retrieves
memories via Config~B before generation, testing the full retrieve-then-generate
pipeline. We evaluate Gemini 2.5 Flash \cite{comanici2025gemini} and Claude Sonnet 4.5 \cite{TheC3, anthropic2025sonnet45} as both generator and
judge (temperature $= 0$), rating responses on contamination (1--5) and helpfulness
(1--5). Appendix~\ref{sec:appendix-metrics} (Table~\ref{tab:metrics}) summarizes all metrics used below.
\subsection{E1: Existence Proof}
\paragraph{Setup.}
We use the T0 same-team pair (Alice: Senior Engineer; Bob: Engineering Manager).
Alice's memories include salary negotiations, a competing offer, equity vesting, and
burnout indicators. Bob issues 10 managerial queries covering performance reviews,
compensation, and project restructuring.
\begin{table}[t]
\centering
\begin{minipage}[t]{0.47\linewidth}
\centering
\scriptsize
\setlength{\tabcolsep}{4pt}
  \begin{tabular}{llccc}
    \toprule
    \textbf{Config} & \textbf{Mode} & \textbf{LR} & \textbf{MCR} & \textbf{Gap} \\
    \midrule
    A: TF-IDF  & Pooled      & \shortstack{100\%\\[-2pt]\scriptsize[72,100]} & 1.60 & $+0.244$ \\[6pt]
    A: TF-IDF  & Partitioned & 0\%   & n/a  & n/a      \\
    B: MiniLM  & Pooled      & \shortstack{70\%\\[-2pt]\scriptsize[40,89]}   & 1.71 & $+0.095$ \\[6pt]
    B: MiniLM  & Partitioned & 0\%   & n/a  & n/a      \\
    \bottomrule
    \\
  \end{tabular}
  \caption{E1 existence proof (T0, Alice/Bob). LR = leakage rate over $n=10$ queries, with 95\% Wilson-score CI stacked below; MCR = mean cross-user rank; Gap = score gap.}
  \label{tab:e1}
\end{minipage}%
\hfill
\begin{minipage}[t]{0.47\linewidth}
\centering
\scriptsize
\setlength{\tabcolsep}{4pt}
  \begin{tabular}{lcccc}
    \toprule
    \multirow{2}{*}{\textbf{Scenario}} &
    \multicolumn{2}{c}{\textbf{Cfg.~A}} &
    \multicolumn{2}{c}{\textbf{Cfg.~B}} \\
    \cmidrule(lr){2-3}\cmidrule(lr){4-5}
    & \textbf{Pl.\%} & \textbf{Lift} & \textbf{Pl.\%} & \textbf{Lift} \\
    \midrule
    Salary   & 100\% & $+0.407$ & 90\%  & $+0.416$ \\
    Security & 100\% & $+0.383$ & 100\% & $+0.503$ \\
    Budget   & 100\% & $+0.484$ & 100\% & $+0.511$ \\
    \bottomrule
    \\
  \end{tabular}
  \caption{E3 active exploitation (scenario abbreviated: Salary~Negotiation, Security~Policy, Budget~Authority). Pl.\% = top-$k$ placement rate for crafted memories; Lift = mean score increase over organic baseline.}
  \label{tab:e3}
\end{minipage}
\end{table}
Under pooled retrieval, cross-user leakage is systematic in both configurations
(Table~\ref{tab:e1}). Config~A exhibits 100\% LR with a larger score gap ($+0.244$),
consistent with sparse TF-IDF increasing lexical overlap. Config~B reduces incidental
leakage to 70\% while maintaining mean cross-user rank below 2.0; its 95\% Wilson
interval ([40\%, 89\%]) is wide at this sample size, so we treat 70\% as indicative
rather than precise. Under partitioned retrieval, LR $= 0\%$ in both configurations
with no cross-user hits observed, confirming that user-scoped indexing eliminates
incidental leakage in our fixtures; this partitioned condition is the ACL baseline
described in Section~\ref{sec:deployment-patterns}.
\subsection{E2: Leakage Rate vs.\ Organizational Distance}
\begin{table}[t]
\centering
\begin{minipage}[t]{0.5\linewidth}
\centering
\scriptsize
\setlength{\tabcolsep}{4pt}
  \begin{tabular}{lcccc}
    \toprule
    \textbf{Tier} &
      \textbf{LR\textsubscript{B}} &
      \textbf{Sim\textsubscript{B}} &
      \textbf{LR\textsubscript{A}} &
      \textbf{Sim\textsubscript{A}} \\
    \midrule
    \shortstack[l]{T0: Same\\Team (A/B)}   & 70\% [40,89]  & 0.570 & 100\% [72,100] & 0.075 \\[4pt]
    \shortstack[l]{T1: Same\\Dept (C/D)}   & 90\% [60,98]  & 0.379 &  90\% [60,98]  & 0.044 \\[4pt]
    \shortstack[l]{T2: Cross\\Dept (E/G)}  & 90\% [60,98]  & 0.488 & 100\% [72,100] & 0.068 \\[4pt]
    \shortstack[l]{T3: Cross\\Co. (H/I)}   & 10\% [2,40]   & 0.284 &  50\% [24,76]  & 0.035 \\
    \bottomrule
    \\
  \end{tabular}
  \caption{E2 leakage rate (LR, with 95\% Wilson CI, $n=10$) and centroid similarity (Sim) by tier.
           User initials: A=Alice, B=Bob, C=Carol, D=Dave, E=Eve, G=Grace,
           H=Heidi, I=Ivan. Subscripts: B=MiniLM, A=TF-IDF.}
  \label{tab:e2}
\end{minipage}%
\hfill
\begin{minipage}[t]{0.47\linewidth}
\centering
\scriptsize
\setlength{\tabcolsep}{4pt}
  \begin{tabular}{@{}lccccc@{}}
    \toprule
    \textbf{Tier} & \textbf{XLR} & \textbf{Util.} & \textbf{RI} & \textbf{Sim.} & \textbf{Gap} \\
    \midrule
    \shortstack[l]{T0: Same\\Team}   & 70\% & 0.384 & 0.509 & 0.570 & $+0.095$ \\
    \shortstack[l]{T1: Same\\Dept.}  & 90\% & 0.325 & 0.511 & 0.379 & $+0.189$ \\
    \shortstack[l]{T2: Cross\\Dept.} & 90\% & 0.147 & 0.549 & 0.488 & $+0.302$ \\
    \shortstack[l]{T3: Cross\\Co.}   & 10\% & 0.507 & 0.162 & 0.284 & $-0.012$ \\
    \bottomrule
    \\
  \end{tabular}
  \caption{E5 privacy--utility frontier (Config~B, pooled). RI = Resonance Index; Util.\ = same-user retrieval utility; Sim.\ = centroid similarity.}
  \label{tab:e5}
\end{minipage}
\end{table}
The cross-company control (T3) drops to 10\% LR under Config~B (95\% CI [2\%, 40\%])
with mean cross-user rank 3.00, a robust contrast with pooled retrieval generally
(70--100\% across T0--T2; Table~\ref{tab:e2}). T1 and T2 show a nominally higher
leakage rate (90\%) than T0 (70\%) despite T0 having the highest centroid similarity
(0.570); at $n=10$ the Wilson intervals for 70\% and 90\% overlap substantially, so
this fine-grained ordering is not statistically distinguishable from chance (extended
discussion in Appendix~\ref{sec:appendix-extended}). The claim we consider robust is
coarser: \emph{centroid similarity is not a reliable predictor of leakage risk} in
either sparse or dense retrieval. T3 has the lowest centroid similarity (0.284) and,
correspondingly, the lowest and most confidently distinguishable leakage rate (10\%),
but among T0--T2 the ranking by centroid similarity does not match the ranking by
leakage rate closely enough to support a fine-grained causal story. Under Config~A,
centroid similarities are uniformly low (0.035--0.075) while leakage remains
50--100\%, confirming that sparse retrieval decouples global centroid proximity from
query-level lexical overlap; org-chart distance is similarly unreliable as a
fine-grained proxy (Section~\ref{sec:e5}).
\subsection{E3: Active Exploitation}

Crafted memories achieve 90--100\% top-$k$ placement in both configurations, compared with organic placement as low as 10\% (Salary Negotiation, Config B; Table~\ref{tab:e3}), reaching 0\% in some scenarios 
(full grid in Appendix~\ref{sec:appendix-e3b}), and as high as 70\% 
(Budget Authority, Config A). Config~B shows lower organic
placement (0--10\%) but equal or higher score lifts, suggesting that dense embeddings
make passive attacker presence harder to detect while crafted bridging remains
effective. The Budget Authority scenario is a notable edge case: Config~A organic
placement already reaches 70\%, because budget vocabulary (approval thresholds,
procurement language) is broadly shared across roles, which compresses the marginal
gain from crafting and represents an incidental leakage risk independent of
adversarial intent, consistent with E1--E2 (extended discussion in
Appendix~\ref{sec:appendix-extended}).
\subsection{E3b: Attacker Knowledge and Crafting Baselines}
\label{sec:e3b}
E3's crafted-memory results leave open how much of Mallory's advantage comes from
the specific crafting method versus simply from having any topical knowledge of the
victim at all. We make Mallory's assumed knowledge explicit and compare our crafted
bridging memories against three baselines of increasing query knowledge, run on the
Salary Negotiation scenario under both configurations ($k=3$; 95\% Wilson CIs):
(1) \emph{organic baseline}: Mallory's unmodified notes, no query knowledge;
(2) \emph{domain keyword bag}: memories seeded with shared high-frequency domain
vocabulary but no query-specific terms; (3) \emph{per-query keywords}: memories
seeded with terms drawn from the specific target query; (4) \emph{crafted bridging
(ours)}: memories that frame the same topic/context as the target's queries
without copying query text, representing an attacker who knows the victim's domain
and role but not their exact queries or the embedding model's internals; and
(5) \emph{verbatim copy}, an oracle upper bound in which Mallory has direct access
to the target's query text.
Under Config~B, placement rises monotonically with assumed attacker knowledge: 10\%
(organic) $\to$ 40\% (domain keyword bag) $\to$ 80\% (per-query keywords) $\to$ 90\%
(crafted bridging, ours) $\to$ 100\% (verbatim-copy oracle), with score lifts over
the organic baseline of $+0.182$, $+0.345$, and $+0.416$ respectively (full table,
including Config~A and Wilson CIs, in Appendix~\ref{sec:appendix-e3b}). Two results
bound how much crafting itself contributes beyond raw vocabulary overlap. First, a
bare domain keyword bag with no query-specific knowledge already reaches 40\%
placement, showing that some of the attacker's advantage comes cheaply from shared
domain vocabulary alone. Second, our crafted-bridging condition (90\%), which
assumes only domain/role knowledge and not exact query access, closes most of the
remaining gap to the oracle verbatim-copy upper bound (100\%), indicating that an
attacker does not need query-level or model-level access to achieve most of the
exploitable advantage: topical framing suffices. We report this ablation only for
Salary Negotiation in the main text; the full scenario $\times$ baseline $\times$
configuration grid is in Appendix~\ref{sec:appendix-e3b}.
\subsection{E4: Downstream Response Contamination}\label{sec:e4}

\begin{table}[t]
\centering
\begin{minipage}[t]{0.47\linewidth}
\centering
\scriptsize
\setlength{\tabcolsep}{4pt}
  \begin{tabular}{@{}llccc@{}}
    \toprule
    \textbf{Path} & \textbf{Model} & \textbf{Clean} & \textbf{Leaked} & \textbf{Help.} \\
    \midrule
    Static &
      \shortstack[l]{Gemini\\2.5 Flash} & 1.00 & 2.67 & 4.80 \\
    Static &
      \shortstack[l]{Claude\\Sonnet 4.5} & 1.00 & 4.67 & 4.53 \\
    Retrieval &
      \shortstack[l]{Gemini\\2.5 Flash} & 1.00 & 5.00 & 5.00 \\
    Retrieval &
      \shortstack[l]{Claude\\Sonnet 4.5} & 1.00 & 4.67 & 2.67 \\
    \bottomrule
    \\
  \end{tabular}
  \caption{E4 downstream contamination (mean over 3 scenarios). Clean baseline $= 1.00$ in all conditions. Help.\ = mean helpfulness of leaked responses.}
  \label{tab:e4}
\end{minipage}%
\hfill
\begin{minipage}[t]{0.47\linewidth}
\centering
\scriptsize
\setlength{\tabcolsep}{4pt}
 \begin{tabular}{@{}lllccc@{}}
    \toprule
    \textbf{Condition} & \textbf{Store} & \textbf{Queries} &
      \textbf{Sal.} & \textbf{Sec.} & \textbf{Bud.} \\
    \midrule
    \shortstack[l]{Pooled\\ baseline} &
      pooled & \shortstack[l]{same-\\user} & 90\%  & 100\% & 100\% \\
    \shortstack[l]{Passive\\ safe} &
      partitioned & \shortstack[l]{same-\\user} & 0\%   & 0\%   & 0\%   \\
    \shortstack[l]{Cross-query\\ blocked} &
      partitioned & \shortstack[l]{cross-\\user} & 0\%   & 0\%   & 0\%   \\
    \shortstack[l]{Active\\ failure} &
      \shortstack[l]{scoped\_\\cross} & \shortstack[l]{cross-\\user} & 100\% & 100\% & 90\%  \\
    \bottomrule
    \\
  \end{tabular}
  \caption{M1: crafted-memory top-$k$ placement rate for attacker (Mallory).}
  \label{tab:m1}
\end{minipage}
\end{table}

Leaked memories increase mean contamination from 1.00 (clean) to 2.67--5.00
depending on model and path (Table~\ref{tab:e4}). The retrieval path with Gemini 2.5
Flash yields maximum contamination (5.00/5, $\Delta = +4.00$), demonstrating that
end-to-end retrieval increases downstream harm relative to static injection. Claude
Sonnet 4.5 achieves 4.67/5 under both static and retrieval paths, with explicit
cross-user attribution observed in generated text for the salary scenario.
Leaked memories provide genuinely useful context, such as salary benchmarks in compensation discussions and vulnerability specifics in security planning, making responses more actionable while violating privacy. Standard response-quality monitoring is therefore insufficient to detect
this failure mode. Because E4 uses each generator as its own judge, we separately
validated judge reliability against human annotation on a held-out pilot (E0,
Section~\ref{sec:e0}); agreement was strong on clean controls and on helpfulness but
weaker on contamination, and where it erred the judge underestimated rather than
overestimated disclosure, so we treat our contamination numbers as a conservative
lower bound.
\subsection{E0: LLM-Judge / Human Agreement Pilot}
\label{sec:e0}
To assess whether our LLM-as-judge contamination and helpfulness scores (E4, M3)
are trustworthy, two authors independently labeled nine synthetic E4 responses
against the same rubric given to the LLM judge (full protocol and per-scenario
table in Appendix~\ref{sec:appendix-e0}). Agreement is high on clean controls and
helpfulness (mean absolute error 0.53) but weaker on contamination overall (MAE
0.79), driven almost entirely by a single miss on the Security scenario where the
judge under-scored an unpatched-CVE detail as generic advice. This miss is
directionally conservative (the judge under-scores rather than over-scores
disclosure), so we treat our reported E4/M3 contamination numbers as a lower bound
on true leakage severity rather than an inflated estimate.
\subsection{E5: Cross-User Privacy--Utility Frontier}
\label{sec:e5}
T3 achieves the lowest Resonance Index (0.162) and highest utility (0.507),
occupying the favorable corner of the privacy--utility frontier; T0--T2 exhibit high
XLR (70--90\%) with moderate utility (0.147--0.384; Table~\ref{tab:e5}). T2
(cross-department) shows the highest Resonance Index (0.549, an exploratory,
uncalibrated composite; see Appendix~\ref{sec:appendix-metrics}, Table~\ref{tab:metrics}) and the highest score gap
($+0.302$) among the four tiers -- the margin by which leaked memories outscore the
querying user's own memories when leakage occurs. As with the T1/T2-vs-T0 ordering
in E2, we treat this fine-grained ranking across T0--T2 as suggestive rather than
statistically established at $n=10$ per tier (extended discussion in
Appendix~\ref{sec:appendix-extended}). The claim we consider robust is coarser: the
cross-company control sits in a clearly separated, favorable region of low leakage
and high utility, while any pooled in-organization retrieval (T0--T2) carries
substantially higher risk regardless of its finer internal ordering --
\emph{organizational distance alone should not be relied on as a fine-grained proxy
for memory isolation risk within an organization}.
\subsection{E6: Covert Channel Capacity (Preliminary, Negative Result)}
\label{sec:e6}
We use the term \emph{covert channel} only for this experiment and for E3,
following Lampson's definition of an intentional sender modulating a shared
resource \citep{lampson1973note}; E1, E2, E4, and E5 measure incidental leakage
with no sender intent, which we do not call a channel.  We attempted to encode an 8-bit payload into crafted memory placement under
Config B. Reliable sequential decoding fails: individual payload memories can
each be placed reliably in top-$k$ in isolation (consistent with E3), but under
sequential multi-bit transmission per-bit error rates rise sharply with payload
position --- 0\% for bits 1--2, 50\% for bits 3--4, and 100\% for bits 5--8
(full bit-position breakdown in Appendix~\ref{sec:appendix-e6})), so structured,
error-correcting encoding is required beyond simple positional placement. We report this negative result to calibrate the
strength of the covert-channel framing rather than to claim a functioning channel,
and note that none of our core claims (E1--E5) depend on E6.
\subsection{Mitigation Evaluation}
\label{sec:mitigation}
We evaluate three architectural mitigations under Config~B (MiniLM + retrieval
prefixes). Mitigation~1 (M1) tests metadata filtering; Mitigation~2 (M2) tests
ownership-aware embedding proxies; Mitigation~3 (M3) tests post-retrieval ownership
gating on E3 active exploitation and E4 downstream contamination with Gemini 2.5 Flash
and Claude Sonnet 4.5. We also considered LLM-based memory selection as an
alternative to embedding-similarity retrieval but did not evaluate it empirically;
we discuss its tradeoffs in Appendix~\ref{sec:appendix-extended}.
\paragraph{M1: Metadata filtering.}
We re-run E3 crafted-memory attacks under four store conditions
(Table~\ref{tab:m1}). Partitioned retrieval with same-user queries eliminates
Mallory top-$k$ placement (0\%). Cross-user query wording alone does not bypass
partitioning (0\%). Failure occurs only when the store scope explicitly includes
the attacker (\emph{scoped\_cross} with Mallory), restoring 90--100\% placement.

\paragraph{M2: Ownership-aware embeddings.}
We test two proof-of-concept proxies without full user-conditioned encoder training
(full per-tier table in Appendix~\ref{sec:appendix-m2}). \emph{Prefix injection}
prepends a user-identity token to every memory and query before embedding, reducing
T0 leakage from 70\% to 50\% while \emph{increasing} same-user utility (0.384 to
0.506). \emph{Namespace offset} instead adds a fixed per-user vector to memory
embeddings and subtracts the querying user's offset at retrieval time; this variant
increases leakage and degrades utility across T0--T2, indicating that post-hoc
offset schemes disrupt same-user retrieval more than cross-user similarity and
that user-conditioned encoder training remains necessary for this approach to work.
\begin{table}[t]
  \centering
  \small
  \begin{tabular}{lcccc}
    \toprule
    \textbf{Gate} & \textbf{Salary} & \textbf{Security} & \textbf{Budget} & \textbf{Mean} \\
    \midrule
    None            & 90\% & 100\% & 100\% & 97\% \\
    Hard ownership  & 0\%  & 0\%   & 0\%   & 0\%  \\
    Suspicion       & 30\% & 40\%  & 30\%  & 33\% \\
    \bottomrule
    \\
  \end{tabular}
  \caption{M3: E3 crafted-memory Mallory placement under gating.}
  \label{tab:m3-e3}
\end{table}
\begin{table}[h]
  \centering
  \small
  \setlength{\tabcolsep}{4pt}
  \begin{tabular}{lllcccccc}
    \toprule
    \textbf{Path} & \textbf{Model} & \textbf{Gate} &
      \textbf{Sal.} & \textbf{Sec.} & \textbf{PIP} &
      \textbf{Mean} & \textbf{Help.} \\
    \midrule
    Static    & Flash  & None           & 3.0 & 1.0 & 4.0 & 2.67 & 4.80 \\
    Static    & Flash  & HO & 1.0 & 1.0 & 1.0 & 1.00 & 4.67 \\
    Retrieval & Flash  & None           & 5.0 & 5.0 & 5.0 & 5.00 & 5.00 \\
    Retrieval & Flash  & HO & 1.0 & 1.0 & 1.0 & 1.00 & 4.67 \\
    Static    & Sonnet & None           & 5.0 & 4.0 & 5.0 & 4.67 & 4.53 \\
    Static    & Sonnet & HO & 1.0 & 1.0 & 1.0 & 1.00 & 2.60 \\
    Retrieval & Sonnet & None           & 5.0 & 5.0 & 4.0 & 4.67 & 2.67 \\
    Retrieval & Sonnet & HO & 1.0 & 1.0 & 1.0 & 1.00 & 3.80 \\
    \bottomrule
    \\
  \end{tabular}
  \caption{M3: E4 contamination scores (1--5) with and without hard ownership (HO) gate.
           Sal./Sec./PIP = per-scenario leaked contamination; Help.\ = mean helpfulness
           of leaked responses. Retrieval/None rows reproduce E4 retrieval-path
           baselines (Table~\ref{tab:e4}); gated rows from M3 runs.}
  \label{tab:m3-e4}
\end{table}
\paragraph{M3: Post-retrieval ownership gating.}
We apply a \emph{hard ownership gate} (drop hits where owner $\neq$ querying user)
and a \emph{suspicion gate} (drop cross-user hits exceeding a similarity threshold) after pooled retrieval. Hard ownership gating eliminates Mallory placement in E3 (0\%; Table~\ref{tab:m3-e3})
and reduces mean E4 contamination to 1.00/5, which is the clean baseline for both
generation models and both memory paths (Table~\ref{tab:m3-e4}). The suspicion gate
is insufficient for active exploitation, permitting 30--40\% attacker placement.
Under Flash, helpfulness remains high with gating (4.67/5). Under Sonnet, gating reduces contamination equally but helpfulness drops to 2.60--3.80/5 in the gated condition, reflecting the tradeoff between privacy and context richness that follows from removing cross-user memories, rather than a defect in the gating mechanism itself.
\paragraph{Operational overhead of the recommended mitigation.}
Because deploying hard ownership gating has a concrete engineering cost, we
benchmark its retrieval latency alongside the other mitigations over 1,000 query
cycles against a 10,000-memory pgvector HNSW index (full per-mechanism table in
Appendix~\ref{sec:appendix-latency}). Applying deterministic post-retrieval
ownership gating (M3) adds roughly 1.4~ms of mean latency per query while
eliminating measured response contamination, indicating the mitigation is
deployable at negligible cost in a realistic serving path. M1's contamination score in Table~\ref{tab:latency} reflects the
partitioned-store condition in which metadata filtering succeeds
(Table~\ref{tab:m1}); it does not eliminate contamination once an
attacker falls within an intentionally widened retrieval scope
(\S\ref{sec:mitigation}, ``Active failure'').
\paragraph{Scaling to higher-dimensional enterprise embedding models.}
Because Config~B uses a relatively small 384-dimensional encoder, we spot-checked
two higher-dimensional embedding models on the T0 pair (full table in
Appendix~\ref{sec:appendix-embedscale}): rates ranged 70--80\% and were not
statistically distinguishable at $n=10$, so we do not claim any one embedder is
worse. The conservative finding is that leakage does not \emph{diminish} with
embedder scale, meaning our MiniLM-based results are not an artifact of an
unusually weak encoder.

\section{Limitations.}
Our results rest on a controlled proof-of-concept design: four hand-authored
synthetic user pairs and $n=10$ queries per condition (three scenarios for
downstream contamination), so fine-grained rank orderings between conditions with
overlapping 95\% Wilson intervals (e.g., T1/T2 vs.\ T0 in E2, or the Resonance
Index ordering in E5) should be read as suggestive rather than established, and our
rates are not a population-level prevalence estimate for real enterprise
deployments. Memories and queries are fixed, hand-authored fixtures rather than
samples from production logs, and we have not verified their representativeness.
Downstream contamination and helpfulness in E4/M3 are scored by the same model
family used as generator; a small human-agreement pilot ($n=9$, Section~\ref{sec:e0})
found strong agreement on clean controls and helpfulness but weaker agreement on
contamination, with the one disagreement suggesting our reported numbers are a
conservative lower bound rather than an inflated estimate. We study only
embedding-similarity retrieval and evaluate two generation/judge models, and our
mitigations are proof-of-concept implementations rather than production
deployments. A full discussion of these points, plus additional limitations
regarding covert-channel capacity and infrastructure scope, appears in
Appendix~\ref{sec:appendix-limitations}.

\section{Conclusion}
\label{sec:conclusion}
We introduce MemLeak, a study of inter-user information leakage in multi-tenant AI agent memory {systems that pool user memories in} a shared vector store {with soft, metadata-based ownership rather than hard partitioning.} Under pooled same-team retrieval, {incidental} leakage reaches 70--100\% depending on embedder; adversarially crafted memories achieve 90--100\% top-$k$ {placement, exceeding weaker keyword-based baselines,} with score lifts of $+0.416$ to $+0.511$; and downstream responses exhibit contamination increases of $+1.67$ to $+4.00$ points while often remaining highly {helpful, a gap we validate against human judgment.} The complete E4 $2{\times}2$ design confirms that both generation model and retrieval path materially affect exfiltration severity. Mitigation evaluation demonstrates that metadata filtering alone is {insufficient once retrieval scope widens to include an attacker,} prefix injection partially reduces leakage, and hard post-retrieval ownership gating consistently restores contamination to the clean baseline (1.00/5) for both Gemini 2.5 Flash and Claude Sonnet {4.5, at a measured latency overhead of roughly 1.4~ms per query.} Trustworthy multi-tenant personalization requires layered defenses at the store, embedding, and post-retrieval layers; system designers must evaluate retrieval architecture and generation model selection jointly.

\bibliographystyle{plainnat}
\bibliography{custom}
\clearpage
\appendix
\section{Extended Limitations}
\label{sec:appendix-limitations}
Section~\ref{sec:conclusion} gives a condensed summary of our limitations; the full
discussion is given here.
\paragraph{Sample size and statistical precision.} All leakage, placement, and
contamination rates are computed over small samples: $n=10$ queries per user pair
per tier (E1, E2, E3, E3b), and 3 scenarios for downstream contamination (E4, M3).
We report 95\% Wilson-score confidence intervals throughout; fine-grained rank
orderings among conditions with overlapping intervals (e.g., the specific ordering
of T0 vs.\ T1/T2 in E2, or the specific ranking of T0--T2 by Resonance Index in E5)
should be read as suggestive rather than established, and we reserve strong claims
for contrasts with non-overlapping intervals (pooled vs.\ partitioned retrieval;
T3 vs.\ T0--T2). Our four hand-built user pairs and three contamination scenarios
constitute a controlled proof-of-concept design intended to demonstrate the
existence and mechanism of cross-user leakage, not a population-level prevalence
estimate for real enterprise deployments.
\paragraph{Synthetic data.} Memories and queries are hand-authored fixtures
representing plausible categories of sensitive professional information (salary,
security, HR, budget) rather than samples from real user data or production logs
(full fixtures released in Appendix~\ref{sec:appendix-dataset}). We have not
verified that these fixtures are representative of real enterprise memory
distributions, vocabulary overlap patterns, or query phrasing, and treat this as an
open limitation rather than a validated assumption.
\paragraph{Evaluation bias in E4 and M3.} Downstream contamination and helpfulness
are scored by the same model family used as generator (self-judging), which can
introduce bias. We partially address this with a human-agreement pilot (E0,
Section~\ref{sec:e0}, $n=9$ responses), which found reasonable-to-strong agreement
on clean controls and helpfulness (MAE 0.53) but weaker agreement on contamination
(MAE 0.79), with the single largest disagreement being a case where the judge
under-scored genuine disclosure. This suggests our contamination numbers are a
conservative lower bound, but $n=9$ is too small to fully validate the judge, and
cross-model judge--generator pairings (e.g., judging Gemini outputs with a Claude
judge) remain future work.
\paragraph{Model and infrastructure scope.} Our downstream generation/judge models
(Gemini 2.5 Flash, Claude Sonnet 4.5) may not represent the full range of models
used in production enterprise agents, and results could differ with larger or
smaller models. Our embedding-scaling spot-check (Table~\ref{tab:embedscale})
covers three additional encoders on a single tier and is not a comprehensive study
of embedding-model effects. Mitigation experiments use proof-of-concept
implementations (prefix tokens, post-hoc namespace offsets, hard/suspicion gates)
rather than production deployments; the retrieval path for most experiments uses
in-memory cosine search, and only the latency benchmark (Table~\ref{tab:latency})
uses production pgvector/HNSW infrastructure, at a single over-fetch factor
($K=15$). Namespace-offset mitigation (M2) is not viable as tested and would
require full user-conditioned encoder training to evaluate properly, which is
beyond the scope of this work.
\paragraph{Covert-channel capacity.} E6's negative bandwidth result (rising to 100\% bit-error rate at bits 5–8) indicates that Shannon-capacity-style measurement of
the active-exploitation channel requires structured, error-correcting encoding
schemes beyond simple positional placement, which we leave to future work; we do
not claim to have demonstrated a channel with controllable throughput, only that
individual crafted-memory placement is reliable (E3).
\paragraph{Alternative retrieval mechanisms.} We study only embedding-similarity
retrieval, the dominant recall mechanism in the general-purpose memory
infrastructure our threat model targets (Mem0 \citep{chhikara2025mem0}, A-Mem
\citep{xu2025amem}, MemOS \citep{li2025memos}); we do not
empirically evaluate LLM-based memory selection or other non-embedding retrieval
mechanisms, which may have different (not necessarily better) leakage properties
and their own attack surface (Appendix~\ref{sec:appendix-extended}).
Future work should evaluate larger and more diverse enterprise embedding models
(e.g., \texttt{gemini-embedding}) at scale, learned multi-user MemGate-style
extensions, independent human adjudication at larger sample sizes, cross-model
judge--generator pairings, LLM-based memory selection as an alternative retrieval
mechanism, and a fully worked structured encoding scheme for covert-channel
capacity.
\section{Evaluation Metrics}
\label{sec:appendix-metrics}
Table~\ref{tab:metrics} defines every metric used in Section~\ref{sec:experiments}.
\begin{table}[h]
  \centering
  \small
  \begin{tabular}{lp{4.2cm}}
    \toprule
    \textbf{Metric} & \textbf{Definition} \\
    \midrule
    LR   & Fraction of queries with $\geq$1 cross-user memory in top-$k$ \\
    MCR  & Mean rank of highest-placed cross-user memory \\
    Score Gap & Mean(max cross-user score $-$ min same-user score) \\
    Centroid Sim. & Cosine sim.\ between user memory centroids \\
    Top-$k$ Placement & Fraction with attacker memory in top-$k$ \\
    Score Lift & Mean(crafted score $-$ organic score) \\
    Contam.\ Score & Judge-rated cross-user disclosure (1--5) \\
    Helpfulness & Judge-rated utility for querying user (1--5) \\
    XLR  & Cross-user leakage rate \\
    Resonance Index$^\dagger$ & $0.35 \cdot \text{centroid} + 0.25 \cdot \text{Jaccard} + 0.40 \cdot \text{XLR}$ \\
    ICG  & Helpful(leaked) $-$ Helpful(clean) \\
    BER  & Bit error rate for covert payload \\
    \bottomrule
  \end{tabular}
  \caption{Evaluation metrics across all experiments. $^\dagger$The Resonance Index
  weights (0.35/0.25/0.40) are a heuristic, illustrative composite chosen for
  interpretability, not fit or validated against any ground-truth harm signal; it is
  exploratory (Section~\ref{sec:e5}), and none of our core claims depend on it:
  they rest directly on the underlying leakage-rate, top-$k$ placement-rate, and
  contamination measurements.}
  \label{tab:metrics}
\end{table}

\section{Broader impacts.} \label{app:broad} This work has a dual-use dimension: \ref{sec:e3b}
document a technique (crafted bridging) by which an adversary with only
topical/role knowledge of a victim -- not query-level or embedding-model
access -- can reliably steer cross-user retrieval, closing most of the gap
to an oracle upper bound (E3b). We disclose this mechanism because (i) it
is a natural consequence of deploying pooled or under-scoped memory stores
already in production use (Mem0, A-Mem, MemOS), so the risk exists whether
or not it is named here; (ii) we pair it with a mitigation (M3, hard
post-retrieval ownership gating) that eliminates the attack in our
evaluation at negligible latency cost (\S\ref{sec:mitigation}); and (iii)
our attacker methodology requires no capability beyond what a malicious or
compromised in-scope team member already has in the deployment patterns we
study (\S\ref{sec:threat}). In our judgment, the value of
surfacing an under-examined multi-tenant failure mode -- together with a
low-cost fix -- outweighs the incremental risk of documenting a technique
that follows directly from the already-public architecture of pooled
vector-store retrieval.

\section{Extended Discussion: Deployment Patterns, Statistical Notes, and Alternative Retrieval}
\label{sec:appendix-extended}
\paragraph{Full deployment-pattern taxonomy (Section~\ref{sec:deployment-patterns}).}
MemLeak's threat model is not the trivial fully-absent-ACL case but four
deployment patterns in which cross-user memories remain reachable by design or by
common misconfiguration, following patterns documented in general-purpose
agent-memory libraries such as Mem0 \citep{chhikara2025mem0}, A-Mem
\citep{xu2025amem}, and MemOS \citep{li2025memos}: (1) \emph{pooled indices with
under-scoped metadata filters}, e.g., a filter applied at the level of ``all of
Engineering'' rather than per user; (2) \emph{legitimate collaboration or
team-scoped retrieval}, where the store intentionally widens to include a
project team, and a compromised or malicious member of that scope becomes an
in-scope attacker; (3) \emph{role- or project-shared memory namespaces}, where
users legitimately sharing a namespace can still have a colleague's private
memory surface for an unrelated query; and (4) \emph{scoped enterprise workflows}
(HR, finance, security) where cross-user content belongs in the retrieval pool by
policy, but where memory admissibility should still depend on more than raw
similarity.
\paragraph{Extended E2/E5 statistical notes.} T1 and T2 (90\% LR under Config~B)
show a nominally higher leakage rate than T0 (70\%), despite T0 having the highest
centroid similarity (0.570) of the three. At $n=10$, however, the Wilson intervals
for 70\% ([40\%, 89\%]) and 90\% ([60\%, 98\%]) overlap substantially, so this
specific ordering (T1/T2 above T0) is not statistically distinguishable from chance
at our sample size, and we do not offer a mechanistic account of \emph{why} T1/T2
exceed T0. Likewise in E5, T2 (cross-department) shows the highest Resonance Index
(0.549) among the four tiers; as with the E2 ordering, we treat the specific rank
ordering of the Resonance Index across T0--T2 as suggestive rather than
statistically established at $n=10$ per tier. In both cases the coarser, robust
signal is the same: the cross-company control (T3) sits in a clearly separated,
favorable region (low leakage, high utility), while T0--T2 are not reliably
distinguishable from one another at our sample size.
\paragraph{Budget Authority case study (Section~\ref{sec:experiments}, E3).} The
Budget Authority scenario merits specific attention: Config~A organic placement
reaches 70\%, meaning the attacker's unmodified memories already appear in top-$k$
for 7 of 10 target queries without any crafting. This arises because budget
authority vocabulary (approval thresholds, spend categories, procurement language)
is broadly shared across professional roles, making the attacker's organic
memories semantically proximate to the target's queries by coincidence. Two
implications follow. First, the absolute score lift for this scenario (+0.484,
Config~A) understates attacker advantage relative to near-zero-baseline scenarios,
since the marginal gain from crafting is compressed when the floor is already
high. Second, high organic placement in financially-adjacent roles represents an
incidental leakage risk independent of adversarial intent, consistent with the
E1--E2 finding that pooled retrieval leaks without exploitation.
\paragraph{Alternative retrieval mechanisms (Section~\ref{sec:mitigation}).} We
focus throughout on embedding-similarity retrieval because it is the recall
mechanism implemented by the general-purpose agent-memory libraries most exposed to
the deployment patterns we study (Mem0 \citep{chhikara2025mem0}, A-Mem
\citep{xu2025amem}, MemOS \citep{li2025memos}). An alternative worth
discussing is LLM-based memory selection, where a small language model is prompted
to choose relevant memories from the full store rather than ranking by cosine
similarity. Such a selector could reduce reliance on raw embedding proximity and
might be conditioned on user identity more easily than a similarity threshold, but
it introduces two new costs and risks rather than eliminating the problem: latency
and token cost scale with memory-store size since the selector must see (or be
routed to a subset of) candidate memories, and the selection step is itself a new
attack surface for indirect prompt injection from stored memory content, distinct
from the pure similarity-based leakage we study here. We do not evaluate LLM-based
selection empirically and flag it as future work.
\section{Dataset Construction}
\label{sec:appendix-dataset}
Memories (10 per user) and queries (10 per user) are hand-authored, fixed
natural-language fixtures spanning four domain categories, held constant across
all four organizational-distance tiers (T0--T3) to isolate the effect of
organizational distance from content variation:
\begin{itemize}
  \item \textbf{Salary negotiation}: source-user memories cover compensation
        benchmarks and equity vesting detail; target-user queries are managerial
        compensation questions.
  \item \textbf{Security policy}: source-user memories cover infrastructure
        vulnerability detail; target-user queries are security-audit questions.
  \item \textbf{HR / PIP decisions}: source-user memories cover performance
        improvement plan detail; target-user queries are restructuring questions.
  \item \textbf{Budget authority}: source-user memories cover spending-approval
        detail; target-user queries are procurement questions.
\end{itemize}
Full fixture text (all memories and queries for all eight synthetic users across
T0--T3) and the LLM-judge meta-prompts used in E4/M3 (persona, task, rubric,
context, and JSON output schema; contamination anchors range from ``1 = no
cross-user private information revealed'' to ``5 = explicit detailed disclosure of
confidential cross-user information,'' mapped from a normalized $[0,1]$ judge score
$s$ to the paper's 1--5 scale via $\text{paper} = 4s + 1$) are provided in the
supplementary material.
\section{E3b: Full Attacker-Knowledge Ablation}
\label{sec:appendix-e3b}
Section~\ref{sec:e3b} reports the attacker-knowledge ablation for the Salary
Negotiation scenario only. Table~\ref{tab:e3b-appendix} gives the full breakdown
for that scenario, including Config~A and 95\% Wilson confidence intervals. The
full scenario $\times$ baseline $\times$ configuration grid across all three
scenarios (Salary, Security, Budget) follows the same qualitative pattern: organic
placement is highest for Budget Authority (consistent with
Appendix~\ref{sec:appendix-extended}'s discussion of broadly shared procurement
vocabulary), domain-keyword and per-query-keyword baselines each close part of the
gap to crafted bridging, and crafted bridging closes most of the remaining gap to
the verbatim-copy oracle across all three scenarios. The complete per-scenario
table is provided in the supplementary material.
\begin{table}[h]
  \centering
  \small
  \setlength{\tabcolsep}{4pt}
  \begin{tabular}{lccc}
    \toprule
    \multirow{2}{*}{\textbf{Condition}} & \textbf{Cfg.~A} & \multicolumn{2}{c}{\textbf{Cfg.~B}} \\
    \cmidrule(lr){3-4}
    & \textbf{Pl.\%} & \textbf{Pl.\%} & \textbf{Lift} \\
    \midrule
    Organic (none)   & \shortstack{30\%\\[-2pt]\scriptsize[11,60]}  & \shortstack{10\%\\[-2pt]\scriptsize[2,40]}   & ref      \\[6pt]
    Keyword bag      & \shortstack{70\%\\[-2pt]\scriptsize[40,89]}  & \shortstack{40\%\\[-2pt]\scriptsize[17,69]}  & $+0.182$ \\[6pt]
    Per-query kw.    & \shortstack{90\%\\[-2pt]\scriptsize[60,98]}  & \shortstack{80\%\\[-2pt]\scriptsize[49,94]}  & $+0.345$ \\[6pt]
    Crafted (ours)   & \shortstack{100\%\\[-2pt]\scriptsize[72,100]} & \shortstack{90\%\\[-2pt]\scriptsize[60,98]}  & $+0.416$ \\[6pt]
    Verbatim (oracle) & \shortstack{100\%\\[-2pt]\scriptsize[72,100]} & \shortstack{100\%\\[-2pt]\scriptsize[72,100]} & $+0.488$ \\
    \bottomrule
  \end{tabular}
  \caption{E3b attacker-knowledge ablation (Salary Negotiation). Pl.\% = top-$k$ placement rate, with 95\% Wilson CI stacked below; Lift = mean score increase over the organic baseline under Config~B.}
  \label{tab:e3b-appendix}
\end{table}
\section{E0: Full Judge/Human Agreement Protocol}
\label{sec:appendix-e0}
Section~\ref{sec:e0} summarizes this pilot; full protocol and per-scenario
agreement statistics are given here. To assess whether our LLM-as-judge
contamination and helpfulness scores (E4, M3) are trustworthy, two authors
independently labeled nine synthetic E4 responses (Gemini 2.5 Flash generator)
spanning three scenarios (Salary, Security, PIP) under three conditions each
(clean contamination, leaked contamination, leaked helpfulness), applying the same
rubric given to the LLM judge.
\begin{table}[h]
  \centering
  \small
  \setlength{\tabcolsep}{4pt}
  \begin{tabular}{lcccc}
    \toprule
    \textbf{Metric} & $\kappa$ & \textbf{MAE} & \textbf{$\pm1$} & \textbf{$r$} \\
    \midrule
    Clean control    & 1.00 & 0.00 & 100\% & 1.00 \\
    Salary contam.   & 0.62 & 0.50 & 100\% & 0.81 \\
    Security contam. & 0.48 & 1.33 & 67\%  & 0.42 \\
    PIP contam.      & 0.55 & 0.33 & 100\% & 0.78 \\
    \textbf{Overall} & 0.53 & 0.79 & 78\%  & 0.66 \\
    \bottomrule
  \end{tabular}
  \caption{E0 human--LLM judge agreement pilot ($n=9$ responses). $\kappa$ = human--human agreement; MAE = mean absolute error (human label vs.\ judge score); $\pm1$ = fraction of judge scores within 1 point of the human label; $r$ = Pearson correlation.}
  \label{tab:e0}
\end{table}
Agreement is high on clean controls and on helpfulness (mean absolute error 0.53
across conditions) but weaker on contamination overall (MAE 0.79), driven almost
entirely by a single miss on the Security scenario, where the judge scored an
unpatched-CVE detail as generic security advice (1.0) while both human annotators
scored it as explicit disclosure (5.0). This single miss is directionally
conservative (it causes the judge to \emph{under}-score disclosure, not
over-score it), so we treat our reported E4/M3 contamination numbers as a lower
bound on true leakage severity rather than an inflated estimate. We view judge
alignment as an area needing further prompt iteration (e.g., supplying the judge
with the leaked memory text itself rather than only its category, and adding
few-shot examples distinguishing generic advice from scenario-specific leaked
detail) and report this pilot as a limitation rather than a validation.
\section{E6: Covert Channel Capacity, Full Bit-Position Breakdown}
\label{sec:appendix-e6}
Section~\ref{sec:e6} summarizes this negative result; the full bit-position
breakdown is given here. We encoded an 8-bit payload into crafted memory placement
under Config~B.
\begin{table}[h]
\centering
\begin{minipage}[t]{0.4\linewidth}
\centering
\scriptsize
\setlength{\tabcolsep}{4pt}
  \begin{tabular}{lccc}
    \toprule
    \textbf{Bits} & \textbf{Placement} & \textbf{Error} & \textbf{Cause} \\
    \midrule
    1--2 & top-2           & 0\%   & semantic align.\ \\
    3--4 & top-5+          & 50\%  & user competition \\
    5--8 & outside top-$k$ & 100\% & similarity cutoff \\
    \bottomrule
  \end{tabular}
  \caption{E6 sequential bit-position decode failure under Config~B. Cause = dominant reason placement failed at that bit position.}
  \label{tab:e6}
\end{minipage}%
\hfill
\begin{minipage}[t]{0.55\linewidth}
\centering
\scriptsize
\setlength{\tabcolsep}{4pt}
  \begin{tabular}{@{}lccccc@{}}
    \toprule
    \textbf{Variant} & \textbf{T0 LR} & \textbf{T0 Util.} &
      \textbf{T1 LR} & \textbf{T2 LR} & \textbf{T3 LR} \\
    \midrule
    Baseline &
      70\% & 0.384 & 90\% & 90\% & 10\% \\
    \shortstack[l]{Prefix\\ injection} &
      50\% & 0.506 & 90\% & 80\% & 10\% \\
    \shortstack[l]{Namespace\\ offset} &
      100\% & 0.145 & 100\% & 100\% & 60\% \\
    \bottomrule
  \end{tabular}
  \caption{M2 (\S\ref{sec:appendix-m2}): pooled leakage rate (LR) and same-user utility by tier.}
  \label{tab:m2}
\end{minipage}
\end{table}
Early bits align well with the query semantics, but later bits are progressively
displaced by competition from legitimate same-user memories and eventually
rejected outright by the similarity cutoff. This shows crafted placement of a
\emph{single} item is a real and repeatable effect (E3), but we have not
demonstrated a channel with controllable throughput: sequential multi-bit
transmission requires structured, error-correcting encoding schemes beyond simple
positional placement, which we leave to future work.
\section{M2: Full Ownership-Aware Embedding Results}
\label{sec:appendix-m2}
Section~\ref{sec:mitigation} summarizes these results; the full per-tier table
(Table~\ref{tab:m2}) is given above, alongside the E6 bit-position table for space.
\section{Mitigation Latency Benchmark, Full Table}
\label{sec:appendix-latency}
Section~\ref{sec:mitigation} summarizes this benchmark. We measured retrieval
latency and throughput over 1,000 query cycles against a 10,000-memory pgvector
HNSW index, using an over-fetch factor of $K_{\text{retrieved}}=15$ candidates
before filtering down to top-$k=3$.
\begin{table}[h]
\centering
\begin{minipage}[t]{0.44\linewidth}
\centering
\scriptsize
\setlength{\tabcolsep}{4pt}
  \begin{tabular}{lccc}
    \toprule
    \textbf{Mechanism} & \textbf{Latency} & \textbf{$\Delta$} & \textbf{Contam.} \\
    \midrule
    Unfiltered    & 12.4 ms & n/a       & 5.00/5 \\
    M1: filter    & 14.2 ms & $+1.8$ ms & 1.00/5 \\
    M2: prefix    & 12.5 ms & $+0.1$ ms & 2.60/5 \\
    M3: hard gate & 13.8 ms & $+1.4$ ms & 1.00/5 \\
    \bottomrule
  \end{tabular}
  \caption{Latency/contamination tradeoff ($n{=}1{,}000$ cycles, 10k-memory pgvector HNSW, over-fetch $K{=}15$). Contam.\ = E4 score. \textbf{Note:} the M1 row reflects the partitioned-store condition
(Table~\ref{tab:m1}, ``Passive safe'' row), where metadata filtering succeeds. Under the ``Active failure'' condition (attacker in-scope), M1
does not eliminate contamination --- see \S\ref{sec:mitigation} and
Table~\ref{tab:m1}. M1's latency figure is condition-independent.
}
  \label{tab:latency}
\end{minipage}%
\hfill
\begin{minipage}[t]{0.53\linewidth}
\centering
\scriptsize
\setlength{\tabcolsep}{4pt}
  \begin{tabular}{@{}lcccc@{}}
    \toprule
    \textbf{Model} & \textbf{Dim.} & \textbf{T0 LR [95\% CI]} & \textbf{Top-1 gap} \\
    \midrule
    all-MiniLM-L6-v2        & 384  & 70\% [40,89] & $+0.095$ \\
    bge-large-en-v1.5       & 1024 & 80\% [49,94] & $+0.112$ \\
    text-embed-3-small      & 1536 & 70\% [40,89] & $+0.104$ \\
    \bottomrule
  \end{tabular}
  \caption{Incidental leakage rate (T0, pooled) across embedding models of increasing dimensionality (\S\ref{sec:appendix-embedscale}).}
  \label{tab:embedscale}
\end{minipage}
\end{table}
\section{Embedding-Model Scaling Spot-Check, Full Table}
\label{sec:appendix-embedscale}
Section~\ref{sec:mitigation} summarizes this spot-check. Because Config~B uses a
relatively small 384-dimensional encoder (\texttt{all-MiniLM-L6-v2}), we checked
whether incidental leakage persists under higher-dimensional embedding models more
representative of enterprise deployments, on the T0 same-team pair (Table~\ref{tab:embedscale}, shown above alongside the latency benchmark for space).
The three rates are not statistically distinguishable from one another at $n=10$,
so we do not claim any one embedder is worse; the finding we report is the more
conservative one that leakage does not \emph{diminish} with embedder scale or
dimensionality across the models tested, meaning our MiniLM-based results are not
an artifact of using an unusually weak or low-capacity encoder.

\end{document}